\documentclass{article}
\usepackage{ijcai22}

\usepackage{times}

\usepackage{soul}
\usepackage{url}
\usepackage[hidelinks]{hyperref}
\usepackage[utf8]{inputenc}
\usepackage[small]{caption}
\usepackage{graphicx}
\usepackage{amsmath}
\usepackage{booktabs}
\usepackage[export]{adjustbox}
\usepackage{multirow}
\usepackage{array}
\usepackage{subfigure}
\usepackage{amssymb}
\usepackage[symbol]{footmisc}
\begin{document}
\title{Entity Alignment For Knowledge Graphs: Progress, Challenges, and Empirical Studies}
%

\author{
Deepak Chaurasiya$^1$$^{\dagger}$\and
Anil Surisetty$^1$$^{\dagger}$\and
Nitish Kumar$^1$\and
Alok Singh$^1$ \and
Vikrant Dey$^{1,2}$\and
Aakarsh Malhotra$^1$\and
Gaurav Dhama$^1$\and
Ankur Arora$^1$\\
\affiliations
$^1$AI Garage, Mastercard India\\
$^2$IIT Roorkee\\
}
%
%
\maketitle              
\begin{abstract}
Entity Alignment (EA) identifies entities across databases that refer to the same entity. Knowledge graph-based embedding methods have recently dominated EA techniques. Such methods map entities to a low-dimension space and align them based on their similarities.
With the corpus of EA methodologies growing rapidly, this paper presents a comprehensive analysis of various existing EA methods, elaborating their applications and limitations. Further, we distinguish the methods based on their underlying algorithms and the information they incorporate to learn entity representations. Based on challenges in industrial datasets, we bring forward $4$ research questions (RQs). These RQs empirically analyse the algorithms from the perspective of \textit{Hubness, Degree distribution, Non-isomorphic neighbourhood,} and \textit{Name bias}.
For Hubness, where one entity turns up as the nearest neighbour of many other entities, we define an $h$-score to quantify its effect on the performance of various algorithms. Additionally, we try to level the playing field for algorithms that rely primarily on name-bias existing in the benchmarking open-source datasets by creating a low name bias dataset. We further create an open-source repository for $14$ embedding-based EA methods and present the analysis for invoking further research motivations in the field of EA.
\end{abstract}
\section{Introduction}
\renewcommand{\thefootnote}{\fnsymbol{footnote}}
\footnotetext[2]{Equal Contribution}
The rise of the internet and multimedia has generated an exponential amount of data. To interpret and use this data efficiently, it needs to be stored in a structured manner. Knowledge Graphs ($\mathcal{KG}s$) \cite{Freebase,YAGO2,google,microsoft} are one such mechanism to store data, leading to its widespread usage. Due to asynchronous data sources, an entity often appears with different aliases in different $\mathcal{KG}s$. This has led to the rise
of EA task, which aims to align entities
from different $\mathcal{KG}s$ alluding to the same real-world entity. Lack of generalization of early methods \cite{el2015alex,hu2011self,heuristics,crowdsourcing} and advancement in $\mathcal{KG}$ representation learning \cite{KRL_1,KRL_2,KRL_3,KRL_4,KRL_5,KRL_6,KRL_7} has led to a rise in embedding based EA methods.

This paper details the working of various existing embedding-based EA methods and compares their applications and limitations.
Traditionally these methods are classified into two classes: (i) Trans-based methods and (ii) GNN-based methods. Trans-based methods learn the representations based on the translation constraint of $head + relation = tail$ \cite{TRANSE}. In contrast, GNN-based methods use Graph Neural Networks to learn structural entity embeddings. 
In this paper, along with the traditional classification, we propose a categorisation based on the type of network information they exploit– (i) Structure aware (S), (ii) Structure and Relation aware (SR), and (iii) Structure and Attribute aware (SA).

This paper complements existing surveys \cite{berrendorf2020knowledge,comprehensive-survey,zhao2020experimental} by providing a comprehensive categorisation of the latest embedding-based EA methods along with inferences based on four RQs, with a focus on EA in the industrial setting. 
An existing study by Sun et al. \cite{OPENEA} provides a survey of EA methods but lacks discussion around the presence of biases in industrial datasets and its effect on models' performance. Our study proposes a novel graph sampling method to generate a low name bias dataset required for rigorous evaluation of methods and provides a metric to quantify the effect of performance adversary phenomena like hubness. Moreover, we provide a comparative evaluation of the latest EA methods specific to the listed RQs to show which methods are better suited for a given industrial setting.
\begin{itemize}
    \item \textbf{RQ1. Does the method address the hubness problem present in \bf{$\mathcal{KG}s$}?} Hubness \cite{hu2011self} in EA is the phenomenon where some entities (hubs) get aligned to many other entities in the vector space. In this study, we propose an $h$-score that measures hubness's extent for an alignment result.
    
\item \textbf{RQ2. Does the method align entities well irrespective of their degree?} The higher the degree of an entity, the higher the structural information present for the same and vice versa. In this paper, we demonstrate the effect of degree on the alignment accuracy of each algorithm.

\item \textbf{RQ3. Does the method align entities having non-isomorphic subgraphs in the two $\mathcal{KG}s$?} Due to distinct sources, an identical entity in two $\mathcal{KG}s$ has non-identical neighbourhoods or, equivalently, non-isomorphic subgraphs. We quantify an algorithm's effectiveness against non-isomorphism by analyzing accurately aligned entities having different 
neighbourhoods.
    
\item \textbf{RQ4. Does name bias lead to an algorithm's high accuracy?} An algorithm's matching result shows name bias when the alignment relies primarily on linguistic information in entities' names. We propose a cross-lingual benchmarking dataset with low name bias to highlight an algorithm's effectiveness.

\end{itemize}

Our major contributions are listed as follows:

\begin{itemize}
    \item A survey of existing methodologies in chronological order and their classification based on the information they capture for EA task. 
    
    \item Extensive experiments to compare the efficacy of existing methods in an industrial setting by drawing empirical inferences based on the listed RQs.

    \item Propose a new graph sampling technique that generates a $\mathcal{KG}$ by sampling entity pairs with low-name bias while keeping graph properties unchanged. Further, we generate and publish a cross-lingual low name-bias dataset for robust testing of EA methods.
    
    \item To ensure reproducibility, we publish the implementation of the listed methods in an open-source library\footnotemark.
\end{itemize}

\footnotetext{Link for the open-source library will be provided post review}

\begin{figure}
    \centering
    \includegraphics[width=0.48\textwidth, height = 3cm, frame]{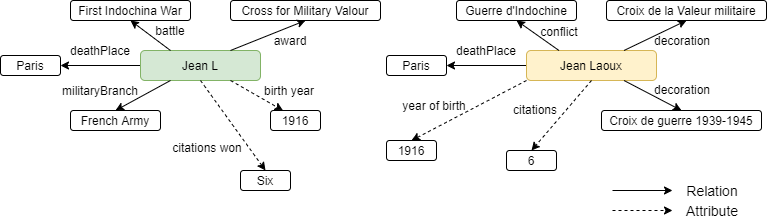}
    \caption{English (left) and French (right) sub-graph for \textit{Jean L}}
    \label{fig:example}
\end{figure}

\begin{figure}
    \centering
    \includegraphics[width=0.48\textwidth, height = 3cm, frame]{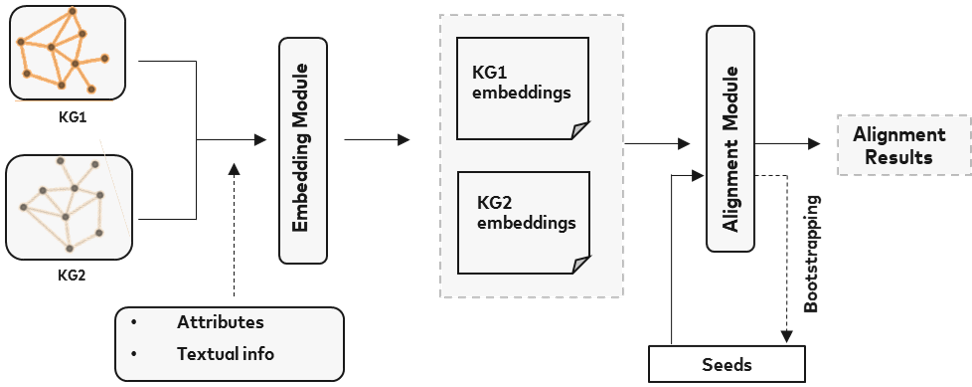}
    \caption{General framework of embedding based EA methods.}
    \label{fig:block_diag}
\end{figure}

\section{Problem Definition}
\label{sec:2}
Here, we introduce preliminary notation that will be consistent for the rest of the paper. A $\mathcal{KG}$ is denoted by $\mathcal{KG}$=($\mathcal{T}$, $\mathcal{A}$, $\mathcal{V}$, $\mathcal{E}$, $\mathcal{R}$),
where $\mathcal{T}$ represents the union of relation and attribute triples, $\mathcal{A}$ represents the set of attributes, and $\mathcal{E}$ represents that of entities. All attribute values and relations form the set $\mathcal{V}$ and $\mathcal{R}$, respectively. General notations used in this study are defined in Table \ref{tab:notations}.

\begin{table}[!htb]
      \centering
      \small
      \caption{Notations}
        \begin{tabular}{|c|c|}
\hline
\textbf{Notation} &\textbf{Description} \\
\hline 
\hline
$h,r,t$& Relation triple (h, t $\in$ $\mathcal{E}$, r $\in$ $\mathcal{R}$)\\
\hline
  $f_{t}(h,r,t)$  & $\|\mathbf{h}+\mathbf{r}-\mathbf{t}\|$\\
  \hline
  $T$ & Set of positive triples i.e $\mathcal{T}_1\cup\mathcal{T}_2$ \\
  \hline
  $T^{\prime}$ & Set of negative triples\\
  \hline
  $h^{\prime},r^{\prime},t^{\prime}$ & A triple $\in T^{\prime}$\\
  \hline
  $l_{tl}(h, r, t)$ &$[\gamma+f_{t}(h, r, t)-f_{t}\left(h^{\prime}, r^{\prime},  t^{\prime}\right)]_{+}$\\
  \hline
    $l_{cl}(h, r, t)$ &$[f_{t}(h, r, t)-\gamma_1]_++[\gamma_2-f_{t}\left(h^{\prime}, r^{\prime},  t^{\prime}\right)]_{+}$\\
  \hline
  $\mathcal{L}_{tl}$ & $\sum_{(h, r, t) \in T}$ $\beta\sum_{h^{\prime},r^{\prime},t^{\prime} \in T^{'}}l_{tl}(h,r,t)$\\
  \hline
  $\mathcal{L}_{cl}$ & $\sum_{(h, r, t) \in T}$ $\beta\sum_{h^{\prime},r^{\prime},t^{\prime} \in T^{'}}l_{cl}(h,r,t)$\\
  \hline
\end{tabular}
    \label{tab:notations}
    \end{table}
    
    \begin{table}
      \centering
        \caption{Classification of EA methods}
        \small
        \begin{tabular}{|m{0.2cm}|m{1.3cm} m{0.7cm} m{0.6cm} m{0.6cm} m{0.8cm} m{0.8cm}|}
            \hline
            & \textbf{Methods} & \textbf{Struc} & \textbf{Rel} & \textbf{Attr} & \textbf{Name} & \textbf{Boot\footnotemark}\\
            \hline
            \hline
            \parbox[t]{2mm}{\multirow{2}{*}{\rotatebox[origin=m]{270}{S}}}
                                             & \multicolumn{1}{l}{GMNN \cite{gmnn}} & \multicolumn{1}{c}{\checkmark} & \multicolumn{1}{c}{-} & \multicolumn{1}{c}{-} & \multicolumn{1}{c}{-} & \multicolumn{1}{c|}{-} \\\cline{2-7}
                                                                              & \multicolumn{1}{l}{AliNet \cite{Alinet}} & \multicolumn{1}{c}{\checkmark} & \multicolumn{1}{c}{-} & \multicolumn{1}{c}{-} & \multicolumn{1}{c}{-} & \multicolumn{1}{c|}{-}\\\cline{2-7}
                             \hline
            \parbox[t]{2mm}{\multirow{9}{*}{\rotatebox[origin=m]{270}{SR}}}
            & \multicolumn{1}{l}{MTransE \cite{MTRANSE}} & \multicolumn{1}{c}{\checkmark} & \multicolumn{1}{c}{\checkmark} & \multicolumn{1}{c}{-} & \multicolumn{1}{c}{-} & \multicolumn{1}{c|}{-}\\\cline{2-7}
            & \multicolumn{1}{l}{IPTransE \cite{IPTRANSE}} & \multicolumn{1}{c}{\checkmark} & \multicolumn{1}{c}{\checkmark} & \multicolumn{1}{c}{-} & \multicolumn{1}{c}{-} &  \multicolumn{1}{c|}{\checkmark}\\\cline{2-7}
            & \multicolumn{1}{l}{BootEA \cite{BOOTEA}} & \multicolumn{1}{c}{\checkmark} & \multicolumn{1}{c}{\checkmark} & \multicolumn{1}{c}{-} & \multicolumn{1}{c}{-} &  \multicolumn{1}{c|}{\checkmark}\\\cline{2-7}
            & \multicolumn{1}{l}{RSN4EA \cite{RSN4EA}} & \multicolumn{1}{c}{\checkmark} & \multicolumn{1}{c}{\checkmark} & \multicolumn{1}{c}{-} & \multicolumn{1}{c}{-} & \multicolumn{1}{c|}{-}\\\cline{2-7}
                                             & \multicolumn{1}{l}{HGCN \cite{hgcn}} & \multicolumn{1}{c}{\checkmark} & \multicolumn{1}{c}{\checkmark} & \multicolumn{1}{c}{-} & \multicolumn{1}{c}{\checkmark} & \multicolumn{1}{c|}{-}\\ \cline{2-7}
                                             & \multicolumn{1}{l}{RDGCN \cite{RDGCN}} & \multicolumn{1}{c}{\checkmark} & \multicolumn{1}{c}{\checkmark} & \multicolumn{1}{c}{-} & \multicolumn{1}{c}{\checkmark} & \multicolumn{1}{c|}{-}\\\cline{2-7}
                                             & \multicolumn{1}{l}{HyperKA \cite{hyperka}} & \multicolumn{1}{c}{\checkmark} & \multicolumn{1}{c}{-} & \multicolumn{1}{c}{-} & \multicolumn{1}{c}{-} & \multicolumn{1}{c|}{-}\\\cline{2-7}
                                             & \multicolumn{1}{l}{MRAEA \cite{MRAEA}} & \multicolumn{1}{c}{\checkmark} & \multicolumn{1}{c}{\checkmark} & \multicolumn{1}{c}{-} & \multicolumn{1}{c}{-} & \multicolumn{1}{c|}{\checkmark}\\\cline{2-7}
                                             & \multicolumn{1}{l}{DualAMN \cite{DualAMN}} & \multicolumn{1}{c}{\checkmark} & \multicolumn{1}{c}{\checkmark} & \multicolumn{1}{c}{-} & \multicolumn{1}{c}{-} & \multicolumn{1}{c|}{\checkmark}\\\cline{2-7}
                                             \hline
             \parbox[t]{2mm}{\multirow{3}{*}{\rotatebox[origin=m]{270}{SA}}}
             & \multicolumn{1}{l}{JAPE \cite{JAPE}} & \multicolumn{1}{c}{\checkmark} & \multicolumn{1}{c}{\checkmark} & \multicolumn{1}{c}{\checkmark} & \multicolumn{1}{c}{-} & \multicolumn{1}{c|}{-}\\\cline{2-7}
             & \multicolumn{1}{l}{AttrE \cite{AttrE}} & \multicolumn{1}{c}{\checkmark} & \multicolumn{1}{c}{\checkmark} & \multicolumn{1}{c}{\checkmark} & \multicolumn{1}{c}{\checkmark} & \multicolumn{1}{c|}{-}\\\cline{2-7}
              & \multicolumn{1}{l}{GCNAlign \cite{gcnalign}} & \multicolumn{1}{c}{\checkmark} & \multicolumn{1}{c}{\checkmark} & \multicolumn{1}{c}{\checkmark} & \multicolumn{1}{c}{-} & \multicolumn{1}{c|}{-}\\ 
              \hline
        \end{tabular}
    \label{tab:categories}
    \end{table} 
\footnotetext{Bootstrapping}
Given two $\mathcal{KG}s$, $\mathcal{KG}_1$ and $\mathcal{KG}_2$, EA aims to identify pairs $(e_1,e_2)$ where 
$e_1 \in \mathcal{E}_1$ and $e_2 \in \mathcal{E}_2$ such that $e_1$ and $e_2$ denote the same entity. Some methods make use of pre-aligned set of entity pairs as training labels which hereby is referred to as seeds. Fig. \ref{fig:example} depicts sub-graphs centered around \textit{Jean\,L} in English and French $\mathcal{KG}$. EA aims to align \textit{Jean\,L} and \textit{Jean\,Laoux} by using knowledge about \textit{Jean} in both the $\mathcal{KG}s$.

\section{Categorisation of Techniques}
\label{sec:4}
For a real-world EA task, it is essential to select an algorithm that optimally captures the information present in the real-world $\mathcal{KG}$. For instance, when entity names are absent in a real-world $\mathcal{KG}$, one would not prefer to use methods like HGCN \cite{hgcn} or RDGCN \cite{RDGCN} since they primarily rely on entity name information for aligning entities. Hence, to ease the process of selecting a suitable method for the problem at hand, we propose categorising all such methodologies based on the information they utilise to align the entities. The methods mentioned above are segregated into $3$ categories: (i) Structure aware (S), (ii) Structure and Relation aware (SR), and (iii) Structure and Attribute aware (SA).
We now describe and compare existing EA methods belonging to each of these categories. These methods use the general framework of an embedding based EA model, as depicted in Fig. \ref{fig:block_diag}.
\subsubsection{Structure aware methods:}
Few popular studies include GMNN \cite{gmnn} and AliNet \cite{Alinet} which use only structure information of $\mathcal{KG}s$ to align entities. GMNN introduced the concept of topic entity matching to EA, which aligns their neighbourhood subgraphs instead of entities.
 Its embedding module uses a 2-layer stacked GCN to encode a node’s neighbourhood structural information, and similar to \cite{GMNN_attn} employs a cross-graph attention mechanism to learn cross-lingual $\mathcal{KG}$ representations. These nodes’ representations are aggregated to generate a topic’s representation, used for final matching. In contrast, AliNet focuses on aligning entities with non-isomorphic neighbourhoods and proposes a multi-hop aggregation embedding module to capture the local and distant neighbourhoods. Its embedding module employs a GCN to generate the local structure representation. To generate distant structure representation, it utilizes an attention mechanism to calculate the contribution of the $n$-hop neighbourhood of an entity. These local and distant representations are aggregated using a gated layer \cite{highway}.
Both methods do not consider relational and attribute information.
\subsubsection{Structure and Relation aware methods:}
This category includes all Trans-based algorithms since they utilise the $head+relation = tail$ constraint while generating entity and relation embeddings. The embedding module of such methods, including MTransE \cite{MTRANSE}, IPTransE \cite{IPTRANSE} and BootEA \cite{BOOTEA} propose variations on the use of the above constraint. MTransE \cite{MTRANSE} embeds the nodes and relations of the two $\mathcal{KG}s$ by optimizing a constraint-based objective i.e.$f_{t}()$ loss, defined in Table \ref{tab:notations}. IPTransE \cite{IPTRANSE}, inspired from PTransE \cite{PTRANSE}, proposed an enhanced learning method by capturing indirect relations between entities using multiple relational paths. It used a triplet-based loss function $\mathcal{L}_{tl}$ where the negative samples are sampled uniformly. The embedding module of BootEA is similar to that of MTransE \cite{BOOTEA}. However, it utilises a contrastive loss function $\mathcal{L}_{cl}$, where the negative samples are generated using $\epsilon$-truncated negative sampling \cite{BOOTEA}. Unlike MTransE, IPTransE and BootEA utilise bootstrapping where entity pairs are aligned during training and appended to the training set iteratively.

Differing from Trans-based methods, Guo et al. \cite{RSN4EA} proposed RSN4EA, which captures long term relation dependencies of entities. Inspired from \cite{node2vec}, RSN4EA generates relational paths by performing biased random walks. These paths are encoded using Recurrent Skipping Network (RSN), with skip connections existing only for entities to distinguish between entities and relations. Guo et al. proposed type based NCE \cite{NCE}, to maximise the likelihood of predicting the following entity in the relational path using the current entity's representation and the learned hidden state for this step.

Superior performance achieved by edge-aware GNNs in graph representation learning tasks has led to the rise of relation aware GNN based EA methods which include HGCN \cite{hgcn}, RDGCN \cite{RDGCN}, HyperKA \cite{hyperka}, MRAEA \cite{MRAEA}, and DualAMN \cite{DualAMN}.
The embedding module of HGCN uses a $2$-layer gated GCN \cite{gcn} to learn the structural embedding of entities, where a relation is represented as the average of associated entities' embeddings.
HGCN represents a node's relational embedding as the average embedding of its associated relations to capture relational information. These structural and relational embeddings are concatenated to generate a node's output representation.
Inspired from \cite{DPGCNN}, RDGCN attempts to jointly learn node and relation embeddings by transforming a $\mathcal{KG}$ (primal) into its dual where each node denotes a particular primal relation, and each edge weight depicts the likelihood of two relations sharing the same head or tail entity in the primal graph. Its embedding module uses a primal-dual attention mechanism, where the primal attention layer captures edge weights using dual graph's node representation and vice versa.
Sun et al. proposed HyperKA \cite{hyperka} to learn representation in a low-dimensional hyperbolic manifold, which enables to capture hierarchies between entities.
HyperKA has a translation based layer followed by $n$-GCN layers, learning representations in a hyperbolic manifold.

To further improve the captured relation information, Mao et al. \cite{MRAEA} proposed MRAEA. MRAEA learns entity and relational representation by capturing relation's semantic properties (direction, type, and inverse). Further, a joint representation of an entity (meta relation aware embedding) is represented as the average of neighbouring entities' representations concatenated with associated relations' average embedding.
The embedding module of MRAEA employs a relation aware attention aggregation mechanism using meta relation aware node embeddings to perform a relation weighted neighbourhood aggregation to generate output embeddings.

Mao et al. further proposed a time-efficient EA method named DualAMN \cite{DualAMN}.
Its embedding uses a simplified relation aggregation layer that utilises a relation reflection operator \cite{RREA}, significantly decreasing the number of parameters required to learn relation-aware representations. Further, DualAMN learns cross $\mathcal{KG}$ interaction by introducing $n$ proxy nodes representing cross-graph alignment relation, which act as reference points for all the nodes in the two $\mathcal{KG}s$. This transforms the learning of node-node cross-$\mathcal{KG}$ interaction problem to learning node-proxy interaction for every node, reducing time complexity from $O(|E_1||E_2|)$ to $O(|E_1|+|E_2|)$. The alignment module of DualAMN utilises LogExpSum \cite{logexpsum1,logexpsum2}, an approximation of the $\varepsilon$-truncated negative sampling technique. LogExpSum enables parallelisation in negative sample generation, improving the overall time complexity.
\subsubsection{Structure and Attribute aware methods:}
The presence of non-isomorphism in $\mathcal{KG}s$ has led to the use of supplementary information, i.e. attributes of entities for EA. Recent studies include JAPE \cite{JAPE}, AttrE \cite{AttrE} and GCNAlign \cite{gcnalign}. Considering that attributes of pre-aligned entities are correlated, JAPE \cite{JAPE} generates attribute representation using a skip-gram \cite{WORD2VEC}.
The objective function aims at maximising attribute co-occurrence.
The embedding module of JAPE is similar to that of MTransE, additionally utilising the attribute similarities for refining the learned embeddings by penalising $L_2$ distance between highly similar entities.

On the other hand, Trisdeya et al. proposed AttrE \cite{AttrE}, an unsupervised EA approach. AttrE treats an attribute triple similar to a relation triple while learning representations. It identifies pseudo aligned entities and relations by employing a string matching algorithm on their names and use these aligned entities as true alignment for its model. The embedding module of AttrE initialises attribute nodes with the machine-translated vector of its value and employs MTransE based embedding module to generate entity, relation and attribute embeddings. GCNAlign \cite{gcnalign} also refrains from differentiating between relation and attribute triples and generates representation for both types of nodes in a unified space. It generates a relation and attribute-based adjacency matrix and pre-processes them to represent each edge by its tf-idf importance.
GCNAlign's embedding module utilises two parallel embedding modules, one for learning structure embeddings and one for learning attribute embeddings. The independent embedding modules are optimised separately, whereas the alignment module uses both structure and attribute embeddings for final alignment.

\section{Implementation}
\label{sec:5}
In our study, we use a total of five datasets to analyze and answer our study's RQs.
These include two sparse cross-lingual datasets: (i) EN-FR$_{V1}$ and (ii) EN-DE$_{V1}$, a dense cross-lingual dataset: (i) EN-FR$_{V2}$, a large cross-lingual dataset: (i) EN-FR$_{100K}$, and a low name bias cross-lingual dataset: (i) EN-FR$_{LNB}$. EN, FR, and DE stand for English, French and German languages, respectively. The first four datasets are from the study Sun et al. \cite{OPENEA} where authors generate the same using Iterative Degree Sampling on DBpedia knowledge base \cite{Dbpedia}, whose statistics are shown in Table \ref{tab:dataset}.
\begin{table*}[hbt]
\centering
\small
\caption{Description of EN-FR$_{V1}$, EN-DE$_{V1}$, EN-FR$_{V2}$, EN-FR$_{100K}$, and EN-FR$_{LNB}$.}
\begin{tabular}{|c|m{1.7cm}| m{1cm}|m{1cm} |m{1cm} |m{1cm} |m{1.75cm}| m{1.9cm}|}
\hline
\textbf{S. no.}&\textbf{Datasets} &\textbf{KGs} &\textbf{\#Ent} &\textbf{\#Rel} &\textbf{\#Attr} &\textbf{\#Rel triples} &\textbf{\#Attr triples}\\
\hline
\hline
1&\multirow{2}{*}{EN-FR$_{V1}$} &EN &$15$K &$267$ &$308$ &$47$,$334$ &$73$,$121$ \\
\cline{3-8}
 &&FR &$15$K &$210$ &$404$ &$40,864$ &$67,167$ \\
\hline
2&\multirow{2}{*}{EN-DE$_{V1}$} &EN &$15$K &$215$ &$286$ &$47$,$676$ &$83$,$755$ \\
\cline{3-8}
& &DE &$15$K &$131$ &$194$ &$50,419$ &$156$,$150$ \\
\hline
3&\multirow{2}{*}{EN-FR$_{V2}$} &EN &$15$K &$193$ &$189$ &$96$,$318$ &$66$,$899$ \\
\cline{3-8}
 &&FR &$15$K &$166$ &$221$ &$80,112$ &$68,779$ \\
\hline
4& \multirow{2}{*}{EN-FR$_{100K}$} &EN &$100$K &$400$ &$466$ &$309$,$607$ &$497$,$729$ \\
\cline{3-8}
 &&FR &$100$K &$300$ &$519$ &$258,285$ &$426,672$ \\
 \hline
5& \multirow{2}{*}{EN-FR$_{LNB}$} &EN &$15$K &$284$ &$273$ &$56$,$088$ &$75$,$805$ \\
\cline{3-8}
 &&FR &$15$K &$223$ &$395$ &$47,720$ &$65,149$ \\
 \hline
\end{tabular}
\label{tab:dataset}
\end{table*} 

\begin{figure}
    \centering
    \subfigure[]{\includegraphics[width=0.49\textwidth, height = 3.5cm]{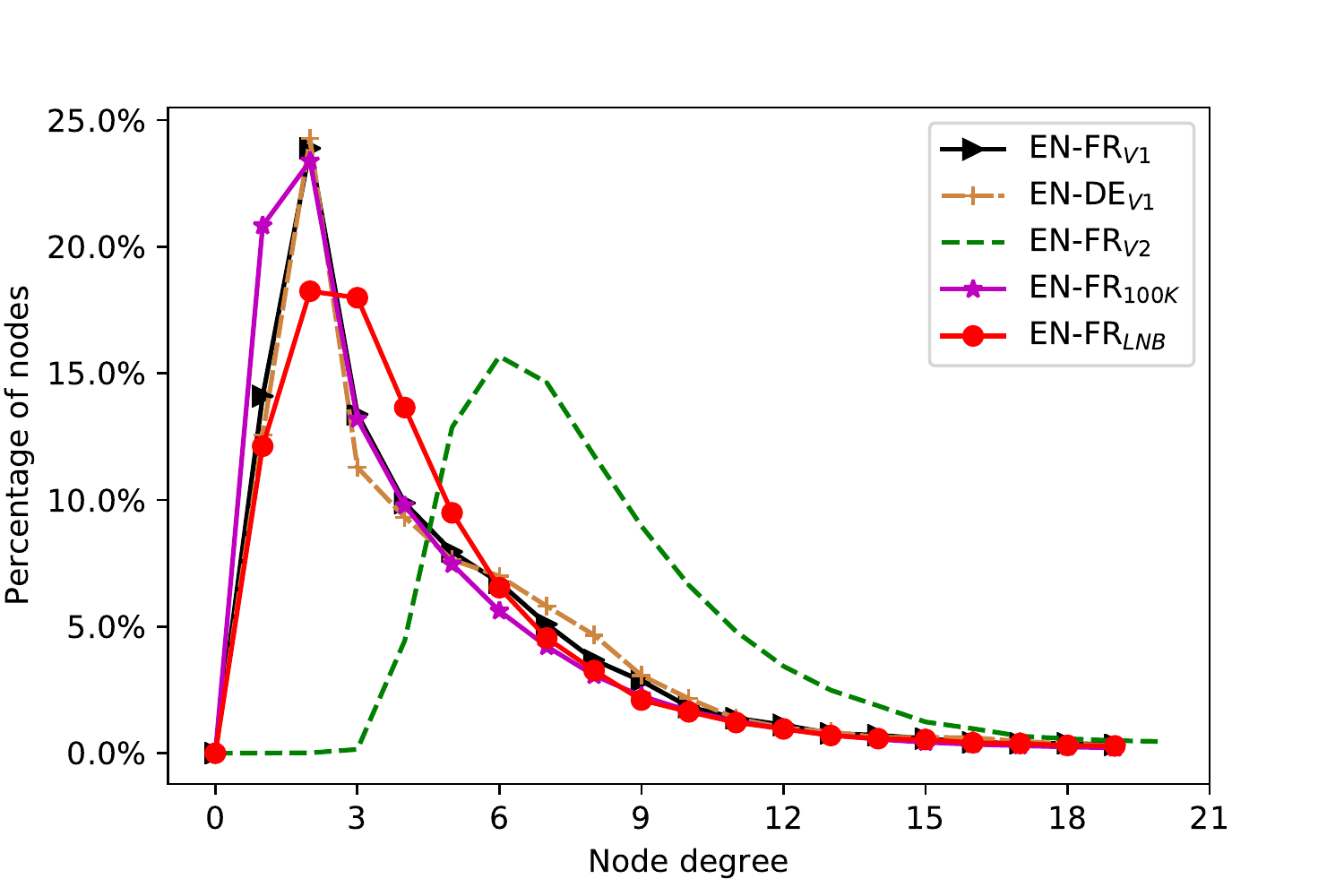}\label{fig:deg_dis}} 
    \subfigure[]{\includegraphics[width=0.5\textwidth, height = 3.5cm]{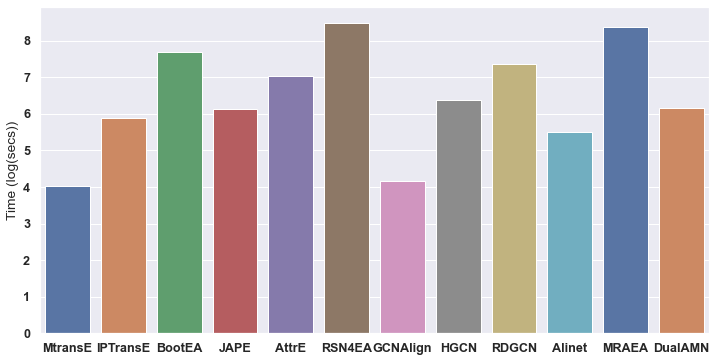}\label{fig:training_time}} 
    \caption{(a) Degree distributions of the five datasets. (b) Training time of existing methodologies on EN-FR$_{V1}$}
    \label{fig:fig34}
\end{figure}

We create the fifth dataset EN-FR$_{LNB}$ (Low Name Bias) to answer RQ4. The entities in EN-FR$_{LNB}$ have relatively low BERT \cite{bert} embedding similarity with their counterparts.
The sampled dataset's degree distribution should be close to source $\mathcal{KG}$ for a fair comparison. Hence, entities are binned based on their degree, and a set of them are sampled uniformly from every bin. The size of the sampled set is proportional to the bin size. To reduce name bias, nodes in the selected set are dropped with a probability proportional to the similarity of its BERT embedding with its counterpart.
Degree distributions of all the datasets are depicted in Fig.\ref{fig:deg_dis}.

All the methods are executed with $20\%$ seeds for training, $10\%$ for validation, and $70\%$ for testing. For a fair comparison, we use a batch size of $5000$ relation triples, maximum training epochs as $1000$, and early stopping on validation Hits@1 with the patience of $10$ epochs for all the methods. For other method-specific hyperparameters such as the number of layers, margins for triplet, and contrastive loss, values proposed by the respective author are followed.
\begin{table*}[hbt!]
\small
\caption{Results of EA methods in terms of Hits@1, Hits@5 and MRR}
\centering
\begin{tabular}{ |c|m{2.1cm}*{12}{|m{0.7cm}}| }
 \hline
 & \textbf{Dataset} & \multicolumn{3}{|c}{\textbf{EN-FR$_{V1}$}} & \multicolumn{3}{|c}{\textbf{EN-DE$_{V1}$}} & \multicolumn{3}{|c}{\textbf{EN-FR$_{V2}$}} & \multicolumn{3}{|c|}{\textbf{EN-FR$_{100K}$}}\\
 \hline
 & \textbf{Method} & H@1 & H@5 & MRR & H@1 & H@5 & MRR & H@1 & H@5 & MRR & H@1 & H@5 & MRR \\
\hline
\hline
\parbox[t]{2mm}{\multirow{2}{*}{\rotatebox[origin=m]{270}{S}}}
&GMNN
\cite{gmnn}
& $\bf74.3$ & $\bf83.8$ & $\bf0.79$ & $\bf83.7$ & $\bf90.7$ & $\bf0.85$ & $\bf80.1$ & $\bf89.7$ & $\bf0.84$ & $\bf68.3$ & $\bf81.5$ & $\bf0.75$\\
&AliNet
\cite{Alinet}
& $36.9$ & $59.3$ & $0.47$ & $60.1$ & $75.8$ & $0.67$ & $54.5$ & $80.5$ & $0.65$ & $33.6$ & $57.1$ & $0.44$\\
\hline
\hline
\parbox[t]{2mm}{\multirow{9}{*}{\rotatebox[origin=m]{270}{SR}}}

& MTransE
\cite{MTRANSE}
& $24.3$ & $45.5$ & $0.34$ & $27.7$ & $49.0$ & $0.38$ & $23.5$ & $42.8$ & $0.33$ & $13.2$ & $25.8$ & $0.20$ \\
&IPTransE
\cite{IPTRANSE}
& $17.4$ & $32.9$ & $0.25$ & $30.5$ & $49.6$ & $0.40$ & $19.9$ & $40.9$ & $0.30$ & $14.2$ & $26.8$ & $0.21$\\
&BootEA
\cite{BOOTEA}
& $50.5$ & $71.6$ &  $0.60$ & $65.5$ & $80.8$ & $0.72$ & $66.1$ & $85.4$ & $0.75$ & $36.1$ & $55.4$ & $0.47$\\
&RSN4EA
\cite{RSN4EA}
& $39.2$ & $59.1$ & $0.48$ & $59.2$ & $75.8$ & $0.67$ & $57.1$ & $74.6$ & $0.65$ & $49.3$ & $66.9$ & $0.57$\\

&HGCN
\cite{hgcn}
& ${\bf77.9}$ & $85.5$ & ${\bf0.81}$ & $82.7$ & $87.5$ & $0.84$ & ${\bf87.5}$ & ${93.1}$ & ${\bf0.90}$ & ${\bf74.3}$ & $77.1$ & ${\bf0.76}$\\
&RDGCN
\cite{RDGCN}
& $75.1$ & $\bf86.4$ & $0.80$ & ${\bf84.1}$ & ${91.5}$ & ${\bf0.87}$ & $81.9$ & $90.9$ & $0.86$ & $71.0$ & $\bf78.2$ & $0.74$\\

&HyperKA
\cite{hyperka}
& $43.5$ & $66.6$ & $0.54$ & $52.2$ & $74.4$ & $0.62$ & $62.2$ & $85.4$ & $0.76$ & $41.8$ & $63.4$ & $0.51$\\
&MRAEA
\cite{MRAEA}
& $53.9$ & $77.2$ & $0.64$ & $70.2$ & $86.6$ & $0.77$ & $74.0$ & $92.4$ & $0.82$
& $38.0$ & $58.8$ & $0.47$\\
&DualAMN
\cite{DualAMN}
& $58.2$ & $78.2$ & $0.67$ & $77.4$ & $\bf92.0$ & $0.83$ & $78.8$ & $\bf94.6$ & $0.85$ & $49.2$ & $68.9$ & $0.58$\\
\hline
\hline
\parbox[t]{2mm}{\multirow{3}{*}{\rotatebox[origin=m]{270}{SA}}}
&JAPE
\cite{JAPE}
& $26.4 $ & $49.3$ & $0.37$ & $29.8$ & $53.3$ & $0.41$ & $30.3$ & $52.7$ & $0.41$ & $30.3$ & $52.7$ & $0.41$\\
&AttrE
\cite{AttrE}
& ${\bf45.2}$ & ${\bf63.6}$  & ${\bf0.54}$ & $48.0$ & $65.9$ & $0.56$ & ${\bf47.0}$ & $69.0$ & ${\bf0.57}$ & ${\bf45.4}$ & ${\bf62.8}$ & ${\bf0.53}$\\
&GCNAlign
\cite{gcnalign}
& $34.4$ & $60.4$ & $0.46$ & ${\bf55.2}$ & ${\bf73.8}$ & ${\bf0.64}$ & $41.9$ & ${\bf70.7}$ & $0.55$ & $26.5$ & $46.2$ & $0.36$\\
\hline
\end{tabular}
\label{tab:results}
\end{table*}
\section{Results}
\label{sec:6}
In this study, we demonstrate our results on $2$ key metrics used in EA- {\bf Hits@k and Mean Reciprocal Rank (MRR)}\cite{MRR}. Table \ref{tab:results} lists the performance of existing methods on the aforementioned benchmarking datasets.

\noindent \textbf{Dense v/s Sparse Dataset}: In general, relation-aware methods perform better on EN-FR$_{V2}$ due to their ability to better incorporate multiple relations associated with every entity. An exception to this is MTransE, where, as the dataset becomes dense, the proportion of entities with multiple relations increases. This results in a drop in performance due to TransE's inability to distinguish entities with similar multi-mapping relations.
The performance of all GNN-based methods also improved due to their ability to capture complex structures around entities.

\noindent \textbf{Training Time}: Fig. \ref{fig:training_time} illustrates the training time of different methods. RSN4EA's training time is relatively high ($86\times$ training time of MTransE) as it trains on a large number of triples created using biased random walks ($5\times$ relation triples in EN-FR$_{V1}$). BootEA's negative sampling and bootstrapping are responsible for its high training time ($39\times$ training time of MTransE). In the case of RDGCN, alternating over the primal-dual graphs is the primary cause for high training time. 
Techniques like MTransE and GCNAlign are the fastest as their embedding modules do not incorporate attention or multi-hop relation triples. Amongst SOTA, DualAMN is the fastest due to fewer parameters, proxy cross graph attention and the use of an approximation of truncated uniform negative sampling. In general, supplementary information such as attributes, multi-hop paths, or techniques like $\varepsilon$-negative sampling boosts the performance but increases the training time simultaneously. 
\section{Analysis}
\label{sec:7}
In this section, we present our analysis towards the four RQs.
\begin{figure*}[h!]
    \centering
    \subfigure[]{\includegraphics[width=0.50\textwidth, height = 3.4cm]{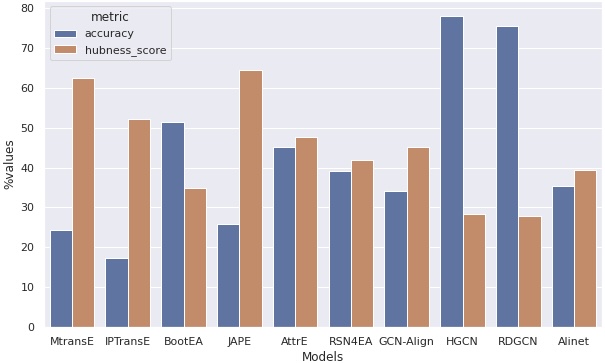}\label{fig:hubness}} 
    \subfigure[]{\includegraphics[width=0.45\textwidth, height = 3.4cm]{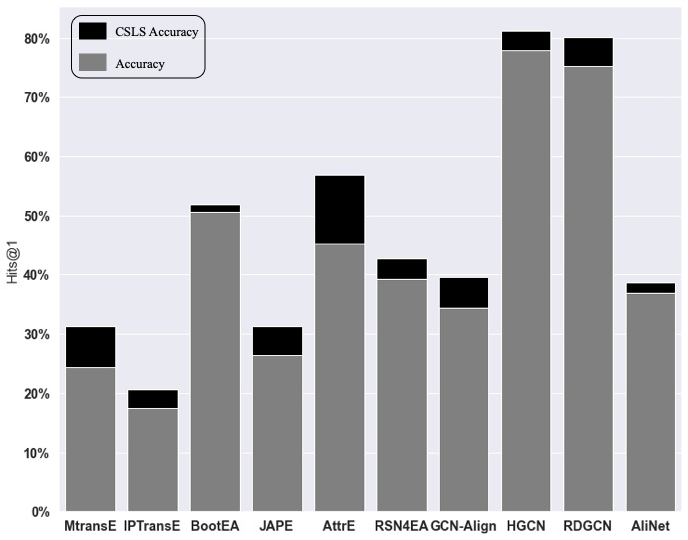} \label{fig:csls_acc}} 
    \subfigure[]{\includegraphics[width=0.52\textwidth, height = 3.3cm]{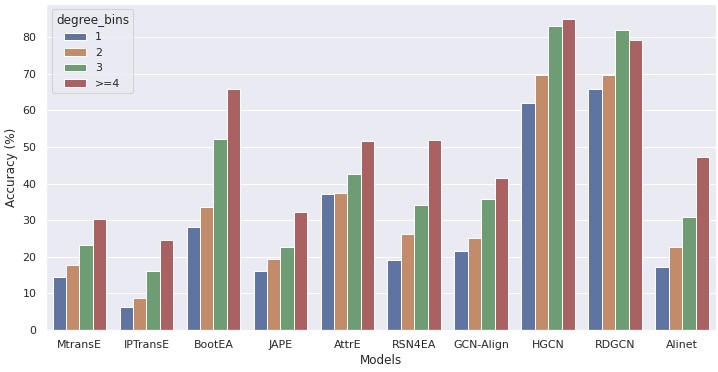} \label{fig:deg}}
    \subfigure[]{\includegraphics[width=0.45\textwidth, height = 3.3cm]{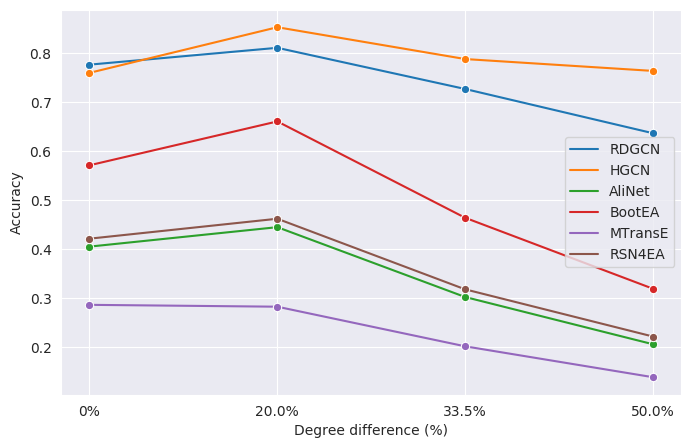} \label{fig:deg_diff}}
    \caption{(a) Hits@1\% and $h$-score on EN-FR$_{V1}$ (b) Hits@1\% with and without CSLS on EN-FR$_{V1}$ (c) Hits@1\% w.r.t. entity degree on EN-FR$_{V1}$ (d) Hits@1\% w.r.t. degree difference on EN-FR$_{V1}$}
    \label{fig:analysis_image}
\end{figure*}
\subsubsection{RQ1. Does the method address hubness problem in $\mathcal{KG}s$?}
Most of the existing methods learn representations in high dimensional vector spaces. Consequently, the curse of dimensionality \cite{curse} leads to the formation of hub nodes \cite{radovanovic2010hubs} that become nearest neighbours of many smaller degree nodes. 
In EA, we observe this phenomenon occurring where an entity gets aligned to many smaller degree entities. For instance, in the alignment results of BootEA on the EN-FR$_{V1}$ dataset, the French entity ``Beillé" gets aligned to $31$ English entities. 
To analyze the effect of hubness in the alignment results, we formulate a $h$-score (hubness score) as:
\begin{equation}
    h\text{-score} = \sum_{z\in \bf{Z}}{hub(z)};\ hub(z) = \frac{\#entities\ aligned\ with\ z}{\#entities\ in\ \mathcal{KG}_i}
\end{equation}
where, $i$ is the $\mathcal{KG}$ to which z belongs to and \textbf{Z} is
the set of largest 10\% hubs (based on $hub()$). A high $h$-score indicates that a higher proportion of entities align with very few hub nodes, which results in low alignment accuracy in injective mapping.
Fig. \ref{fig:hubness} lists the prediction accuracy and $h$-score of various existing EA methods.
Trans-family EA algorithms, such as MtransE and JAPE, rely only on structural and relational information and have a high $h$-score, which shows a pronounced effect of hubness. Whereas algorithms that use external information (RDGCN, HGCN) decrease the effect of hubness due to the use of lingual information to differentiate an actual and a potential match (hubs). To overcome hubness, Conneau et al. \cite{csls} proposed Cross-domain Similarity Local Scaling (CSLS). CSLS aims to decrease the similarity associated with vectors lying in dense areas and is effective against hubness. The same can be observed in Fig. \ref{fig:csls_acc}. To conclude, external information and modifications in the alignment strategy like CSLS can help suppress the effect of hubness in EA, consequently improving the overall accuracy.
\subsubsection{RQ2. Does the method align the entities well irrespective of their degree?}
Degree distribution of a $\mathcal{KG}$, shown in Fig. \ref{fig:deg_dis}, reflects that $~50\%$ of the entities have degree $<4$. This negatively skewed distribution indicates that most entities have a low degree, implying less structural information to learn the embeddings. For a highly connected entity, many neighbouring entities influence their structural information, resulting in generalized representation, and for low degree entities, the representation is sub-optimal as only a few neighbours influence it.

Fig. \ref{fig:deg} plots the alignment accuracy of algorithms for entities with a specific degree. The general trend reaffirms our hypothesis that higher degree entities have better alignment accuracy. Also, we observe that algorithms that use supplementary information of entities show a marginal difference in alignment accuracy of smaller and larger degree nodes. These include RDGCN and AttrE.
Hence, we conclude that algorithms primarily relying on the structure can lead to sub-optimal performance for very sparse datasets. The use of node attributes and linguistic information to supplement structure can improve alignment results.

\subsubsection{RQ3. Does the method align entities having non-isomorphic neighbourhood in the two $\mathcal{KG}s$?}
$\mathcal{KG}s$ are usually incomplete \cite{incompletness}, leading to different degrees of identical entities in the two $\mathcal{KG}s$ resulting in a non-isomorphic neighbourhood. We study the effectiveness of existing methods in aligning entities with different degrees. A negative degree difference would imply that a less connected entity in $\mathcal{KG}_1$ appears as a highly connected entity in $\mathcal{KG}_2$ and vice-versa. 

Fig.\ref{fig:deg_diff} shows how accuracy varies for different algorithms when aligning nodes with the non-isomorphic neighbourhood. RDGCN and HGCN perform degree-difference invariant alignment as they primarily utilize the name linguistic similarity of entities. 
Further, AliNet’s performance varies marginally with increasing non-isomorphism as the former uses attribute predicates and structural information while the latter looks at the multi-hop neighbourhood of an entity.
Contrary to them, MTransE shows a drop in alignment accuracy, which reassures our hypothesis that algorithms relying solely on structural information struggle to align non-isomorphic entities. Hence, the use of complementary information about entities such as attribute triples, lingual information, or global network properties, along with structural information, is essential in handling non-isomorphic subgraphs.

\begin{table}[!htb]
      \centering
      \small
      \caption{\centering Hits@1\% on EN-FR$_{V1}$ $\&$ EN-FR$_{LNB}$}
        \begin{tabular}{|c|c|c|c|}
    \hline
         & \textbf{AttrE} & \textbf{HGCN} & \textbf{RDGCN} \\
         \hline
         \hline
    EN-FR$_{V1}$ & 45.25 &77.93 & 75.17\\
    \hline
    EN-FR$_{LNB}$ & 38.77& 65.63 & 61.10\\
    \hline
    Difference &6.48 & 12.30 & 14.07 \\
    \hline
    \end{tabular}
    \label{tab:name_bias)}
    \end{table}%

\subsubsection{RQ4. Does name-bias in the data lead to an algorithm’s high accuracy?}
Techniques that use textual information cannot be adjudged with other algorithms because of high similarities of linguistic information with counterparts. For instance, nearest neighbour search using BERT embeddings on EN-FR$_{V1}$ gives Hits@1 of $0.69$, indicating a very high name bias. Table \ref{tab:name_bias)} quantifies the sensitivity of EA techniques towards entity names. It compares the results on EN-FR$_{V1}$ dataset against a low name biased EN-FR$_{LNB}$ dataset having $0.57$ Hits@1 with BERT embedding. The significant dip in the performance of RDGCN and HGCN reflects their strong dependency on entity names.
We believe that these methods are unsuited for scenarios where the name similarity of entities in the two complementary $\mathcal{KG}s$ does not bear any significance. For instance, consider the task of aligning debit cards, identified by their number, which belongs to the same cardholder. Such algorithms struggle in this scenario as the two debit card identifiers contains no linguistic similarity. Contrarily, AttrE sees a relatively less drop in performance as it uses multiple attribute information along with entity names. Therefore, we conclude that for industrial datasets with encryptions or low-name bias, methods such as AttrE should be preferred.

\section{Conclusion}
\label{sec:8}
In this paper, we present a comprehensive overview of the various methods and techniques in the field of entity alignment between $\mathcal{KG}s$ and conduct a benchmarking study of the representative approaches. We investigate the existing methods on four research questions inspired by limitations in real-world datasets. We propose a novel graph sampling technique to ensure robust testing of methods to sample a low name-bias dataset. Further, we develop an open-source library with multiple EA approaches and publish their implementations along with the aforementioned low-name bias dataset. This study provides a more profound understanding of EA techniques and the effect of different experimental settings on their performance. The challenges and inferences presented in this work pave the path for future direction for the research community.

\bibliographystyle{splncs04}
\bibliography{pakdd}
\end{document}